\documentclass[10pt,twocolumn,letterpaper]{article}
\usepackage{algorithm}
\usepackage{algpseudocode}
\usepackage{multirow}
\usepackage{amsmath}
\usepackage{mathtools}
\usepackage{longtable}
\usepackage{authblk}
\usepackage{booktabs} 
\usepackage[accsupp]{axessibility} 
\usepackage[pagenumbers]{cvpr} 

\definecolor{cvprblue}{rgb}{0.21,0.49,0.74}
\usepackage[pagebackref,breaklinks,colorlinks,allcolors=cvprblue]{hyperref}

\def\paperID{11} 
\def\confName{CVPR}
\def\confYear{2026}

\title{Data-Free Contribution Estimation in Federated Learning \\ using Gradient von Neumann Entropy} 

\makeatletter
\renewcommand\AB@affilsepx{\qquad \protect\Affilfont}
\makeatother

\author[1]{Asim Ukaye\thanks{Corresponding author: asim.ukaye@mbzuai.ac.ae}}
\author[1]{Mubarak Abdu-Aguye}
\author[1]{Nurbek Tastan}
\author[1,2]{Karthik Nandakumar} 

\affil[1]{MBZUAI, UAE}
\affil[2]{Michigan State University, USA}

\begin{document}
\maketitle
\begin{abstract}
Client contribution estimation in Federated Learning is necessary for identifying clients' importance and for providing fair rewards. Current methods often rely on server-side validation data or self-reported client information, which can compromise privacy or be susceptible to manipulation.
We introduce a data-free signal based on the matrix \text{ von Neumann (spectral) entropy}  of the final-layer updates, which measures the diversity of the information contributed. We instantiate two practical schemes: (i) \emph{SpectralFed}, which uses normalized entropy as aggregation weights, and (ii) \emph{SpectralFuse}, which fuses entropy with class-specific alignment via a rank-adaptive Kalman filter for per-round stability.
Across CIFAR-10/100 and the naturally partitioned FEMNIST and FedISIC benchmarks, entropy-derived scores show a consistently high correlation with standalone client accuracy under diverse non-IID regimes -- without validation data or client metadata. 
We compare our results with data-free contribution estimation baselines and show that spectral entropy serves as a useful indicator of client contribution.
The code is available at: \href{https://github.com/asimukaye/spectralfuse}{\color{magenta} github.com/asimukaye/spectralfuse}
\end{abstract}
\vspace{-1em}
\section{Introduction}
\label{sec:intro}

Federated Learning (FL) enables the training of machine learning models across a population of clients without directly sharing their private data \cite{mcmahanCommunicationEfficientLearningDeep2017}. This paradigm addresses fundamental concerns about data privacy and regulatory compliance while leveraging distributed computational resources. Despite its promise, FL introduces unique challenges beyond standard centralized training. Chief among these are the statistical heterogeneity of client data, system heterogeneity of client devices, and the need for mechanisms to ensure trustworthy participation. \cite{kairouz2021advances}.

\noindent A central yet underexplored aspect of trustworthiness in FL is the estimation of each client's contribution to the global model. Existing aggregation schemes, such as FedAvg \cite{mcmahanCommunicationEfficientLearningDeep2017}, implicitly assume honest self-reporting (e.g., sample counts), while other approaches, such as CFFL \cite{lyu2020collaborative} and FedCE \cite{jiang2023fair}, rely on an auxiliary validation set to measure client model performance. Both assumptions can be problematic in practice: validation sets may not exist or may be biased relative to the clients' distributions, and self-reported metadata can be strategically misrepresented, especially in incentive-based settings. Recent data-free approaches attempt to circumvent these issues by evaluating client updates without direct access to private data. For instance, CGSV \cite{xuGradientDrivenRewards2021} scores clients by the cosine similarity between their gradients and the cohort average, yet such similarity-based measures may penalize genuinely informative but dissimilar updates. Authors of ShapFed \cite{tastan2024redefining} propose Class-specific Shapley Values (CSSV) that computes class-wise alignment of the final layer of the model update. They show that this approach helps in improving the contribution estimates by accounting for label skew present in real-world data settings.

\begin{figure*}
    \centering
    \includegraphics[width=\linewidth]{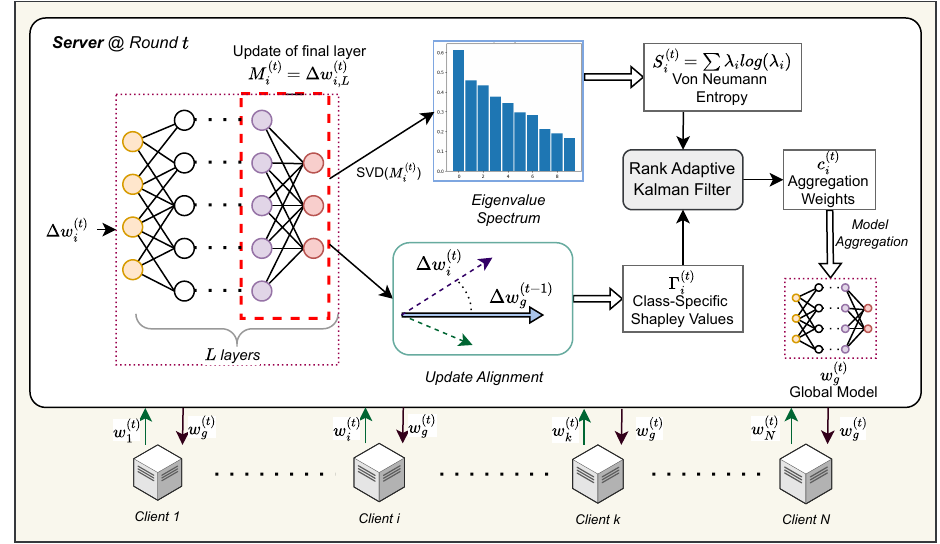}
    \caption{Illustration of the proposed SpectralFuse approach. We incorporate the von Neumann Entropy and Class-specific Shapley values of the clients final-layer updates to form a robust client contribution estimate which serves as  weights for global model aggregation.} 
    \label{fig:main}
\end{figure*}

\noindent In this work, we propose a novel \emph{data-free, non-self-reported} metric for estimating client contribution in FL based on the \textbf{matrix von Neumann entropy} or spectral entropy \cite{benzi2023computation}, \cite{peres2002quantum} of each client's model update. Intuitively, the spectral entropy of the final layer’s weight update captures the diversity and richness of a client’s learned representation without requiring access to any data. We show that higher entropy correlates with more informative client updates and thus greater contribution to the global model.

\noindent While spectral entropy is a strong per-round indicator, the true contribution of a client drifts over rounds due to local learning progress. Per-round measurements are therefore noisy and occasionally volatile. To turn these noisy observations into stable scores, we introduce a \emph{Rank-Adaptive Kalman Filter} \cite{kalmanfilter} component that models a client’s latent contribution as a hidden state evolving over time and treats spectral entropy as one of its observations. It uses CSSV as another observation and fuses them together to obtain a robust contribution estimate.
We propose two variants of our approach: \textbf{SpectralFed}, which uses normalized spectral entropy directly for aggregation weights, and \textbf{SpectralFuse}, which fuses spectral entropy with class-specific Shapley values over time via a Kalman filter update to produce robust contribution estimates. 

\noindent We conduct extensive experiments on CIFAR-10 and CIFAR-100 under multiple partitioning regimes using different model architectures, as well as on the naturally partitioned FEMNIST \cite{caldas2018leaf}, and FedISIC \cite{ogierduterrailFLambyDatasetsBenchmarks2022} datasets. For each client, we compute entropy-derived weights (for Spectralfed) and fusion-derived weights (for SpectralFuse) and compare these to standalone client accuracies. Our results show a strong positive correlation between the proposed weights and client standalone performance across most settings, suggesting that the metric reliably reflects client quality. When incorporated into the aggregation process, our weighting scheme yields global models with test accuracies that are on par with or outperform standard baselines.


\noindent The key contributions of this paper are:
\begin{itemize}
\item We introduce a \emph{data-free, non-self-reported} contribution signal using the von Neumann entropy of client updates.
\item We formulate a \emph{Rank Adaptive Kalman Filter} that treats entropy and CSSV as noisy observations of a latent contribution-estimation process.
\item We instantiate these ideas in \textbf{SpectralFed} (direct entropy weighting) and \textbf{SpectralFuse} (fusion-based weighting), integrating them into standard FL process.
\item We provide an extensive evaluation across vision benchmarks under non-IID regimes, showing strong correlation with client quality and competitive predictive performance, without the need for auxiliary data or metadata.
\end{itemize}


\section{Related Work}
\label{sec:rel}

\noindent\textbf{Client Contribution Estimation in Federated Learning.} 
A central challenge in FL is to fairly weight or reward clients based on their actual contribution to the global model. Early work such as FedAvg \cite{mcmahanCommunicationEfficientLearningDeep2017} uses the number of local samples as a proxy for client weight, implicitly assuming honest self-reporting. However, when incentives are tied to contribution, this assumption may not hold. More recent approaches rely on an auxiliary validation set to assess each client’s update. For example, CFFL \cite{lyu2020collaborative} and FedCE \cite{jiang2023fair} evaluate client models on server-side validation data to estimate their goodness. Yet, validation sets may be unavailable or biased relative to client data distributions, limiting the reliability of these metrics.

\noindent\textbf{Data-Free and Non-Self-Reported Methods.}  
To avoid privacy risks or reliance on client metadata, several works propose data-free scoring of client updates \cite{xuGradientDrivenRewards2021, tastan2024redefining, tastan2025cycle, tastan2025aequa}. CGSV \cite{xuGradientDrivenRewards2021} computes the cosine similarity between the gradient of each client and the average gradient of the cohort, rewarding updates that are better aligned with the average of the group. ShapFed \cite{tastan2024redefining} improves upon the gradient alignment strategy by proposing class-wise gradient alignment for the final layer of the model. This allows for better estimates of client contributions in scenarios with high skew in class labels within each client. Fed-LWR \cite{yan2024new} introduces layer-wise relevance weighting of client updates, while Fed-PCA \cite{wu2025fedpca} leverages pairwise correlated agreement to score clients without a test set and uses those scores for aggregation and a strategy-proof incentive mechanism. FedBary \cite{liDataValuationDetections} computes a Wasserstein barycenter over client distributions and uses barycentric distances as validation-free relevance scores. These methods highlight a growing interest in server-side evaluation of client updates without raw data. 

\noindent Despite their promise, similarity-based or projection-based methods can mischaracterize genuinely useful but diverse updates. Moreover, Shapley-based approaches often incur high computational costs and scale poorly with client count.

\noindent\textbf{Weight-based Measures.}  
A growing line of work gauges model quality directly from weight-space geometry, requiring no training or test data. An initial empirical analysis shows that several training-free or minimally supervised proxies, grounded in weight spectra, are competitive predictors of generalization \cite{jiang_fantastic_2019}. Random-matrix analyses argue that deep networks exhibit implicit self-regularization, where heavy-tailed spectra and scale metrics track generalization \cite{martin2021implicit}. Building on this perspective, \emph{WeightWatcher} predicts the accuracy trends of state-of-the-art networks using power-law exponents and other spectral properties without data \cite{martinPredictingTrendsQuality2021}.
Similar conclusions are shown in NLP, where training-free generalization metrics computed from parameters correlate with downstream performance across tasks \cite{yang_evaluating_2023}.  
Further, spectral or information-theoretic measures have been used to characterize model complexity or representation diversity \cite{benzi2023computation}. However, to our knowledge, von Neumann entropy of client weight updates has not yet been explored as a metric for contribution in FL. 


\noindent In summary, our work differs from prior methods by (i) requiring no auxiliary validation data, (ii) avoiding similarity-to-average assumptions, and (iii) offering an inexpensive, information-theoretic proxy for client contribution that correlates with standalone performance across diverse datasets.

\section{Preliminaries} 
\label{sec:prelim}

We consider a standard cross-silo federated learning scenario with $N$ clients indexed by $i \in \mathcal{K}$, where $\mathcal{K} = \{1, 2, \ldots, n\}$.  The optimization problem posed for federated learning is to minimize the objective:
\begin{align}
    \label{eq:opt}
    f^{\star} = \min_{w \in \mathbb{R}^d} f(w)  ,& \quad\text{where} \quad  f(w) \coloneqq \frac{1}{n} \sum_{i=1}^n f_i(w) \\ 
    \ f_i(w) &\coloneqq \mathbb{E}_{z_i\sim \mathcal{D}_i}\big[F_i(w, z_i)\big] 
\end{align}
\noindent where $F_i$ is typically the loss of the prediction over sample set $z_i$ drawn from the client distribution $D_i$. In real-world settings, the client data distributions may not be identically distributed, i.e. $D_i \neq D_j$, which leads to inherent data heterogeneity in the client cohort. 
In a typical federated training setup, the clients and server communicate their models for a total of $T$ communication rounds.
At communication round $t$, the server broadcasts the global model parameters $w_g^{(t)}$ to all participating clients who set this as their initial model $w_{g|t-1}^{(t)}$ . The clients then update their models using SGD and send back the updated model $w_i^{(t)}$ to the server. The server computes a new global model using an aggregation scheme , $w_g^{(t+1)}= G(w_i^{(t)}, \mathbf{c}, \ldots)$, where $\mathbf{c}$ is an aggregation coefficient used to perform optimal model aggregation. Different approaches use different variants of $\mathbf{c}$ along with additional parameters to obtain the global model. We focus on the subclass of methods, where $\mathbf{c} \in \mathbb{R}^N$ is the weight vector consisting of each client's weight as a scalar used for the linear weighted aggregation of the client models.
Our goal is to compute an online \emph{data-free contribution estimate}, 
$\mathbf{c}^{(t)} = [c_i^{(t)}]^T$, that reflects the client’s true contribution to the global model. To motivate this better, we first look at the standard Federated Averaging algorithm.

\paragraph{Federated averaging.} A common approach to solving (\ref{eq:opt}) is FedAvg \cite{mcmahanCommunicationEfficientLearningDeep2017}. This algorithm involves the participants performing several local steps of SGD  and communicating with the server over multiple communication rounds. In each communication round, $t$, the updates from the participants are averaged on the server and sent back to all participants. For a local epoch $e$ and participant $i \in \mathcal{K}$, the local iterate is updated according to: 
\begin{equation}
    w_g^{t+1} = \sum_{i=1}^n \frac{m_i}{M} w_i^t.
\end{equation} 
where, $m_i$ is the total data size held by a client and $M = \sum_{i=1}^n m_i$ is the total data size across all clients. Note that $m_i$ is a self-reported metric by the client. A variant of the FedAvg algorithm, FedOpt \cite{reddiAdaptiveFederatedOptimization2021}, uses a server side optimizer to perform aggregation. 


\noindent\textbf{Von Neumann Entropy.} 
Given a symmetric positive semi-definite matrix $M$ with eigenvalues $\{\lambda_j\}$, the von Neumann entropy is defined as:
\begin{equation}
S(M) = -\sum_{j}\lambda_j \log \lambda_j.
\label{eq:entropy_definition}
\end{equation}
Intuitively, this quantity measures the “spectral spread’’ or information content of $M$. In other words, matrices with more uniform eigenvalue distributions have higher entropy.\\

\noindent\textbf{Kalman Filter.}
Kalman Filters are a popular choice of statistical filters used in the sensor fusion and control theory domains. They are known to be statistically optimal under Gaussian noise assumptions \cite{kalmanfilter}. 
The typical Kalman Filter is described as follows.  
Given $x_{i}^{(t)}$ as the target state to be observed at time $t$, and $y_{i}^{(t)}$ as the measurements for that state.
A standard Kalman Filter update is then given as:\\
\noindent \textbf{Predict:}
\begin{equation}
    \hat{x}_{i|t-1} = \hat{x}_{i}^{(t-1)}, 
\quad
P_{i|t-1}=P_{i}^{(t-1)}+Q, \label{eq:kf_pred} \\
\end{equation}
\noindent \textbf{Update:}
\begin{align} \label{eq:kf_upd_1}
K_{i}^{(t)}&=P_{i|t-1}H^{\!\top}\!
\Bigl(H\,P_{i|t-1}H^{\!\top}\! + R_{i}^{(t)}\Bigr)^{-1},\\
\hat{x}_{i}^{(t)}&=\hat{x}_{i|t-1}+K_{i}^{(t)}
\Bigl(y_{i}^{(t)}-H\hat{x}_{i|t-1}\Bigr),\label{eq:kf_upd}\\
P_{i}^{(t)}&=(I-K_{i}^{(t)}H)P_{i|t-1}.
\end{align}
\noindent where, $H$ is the mapping from the measurement space to state space. $P_{i}^{(t)}$ is the state covariance, $K_{i}^{(t)}$ is the Kalman Gain , and $Q$ and $R$ are the process and measurement noise covariances, respectively.
We refer the interested reader to \cite{kalmanfilter} for further details on the Kalman Filter.

\section{Methodology}

\label{sec:method}

\begin{figure}
    \centering

    \includegraphics[width=\linewidth]{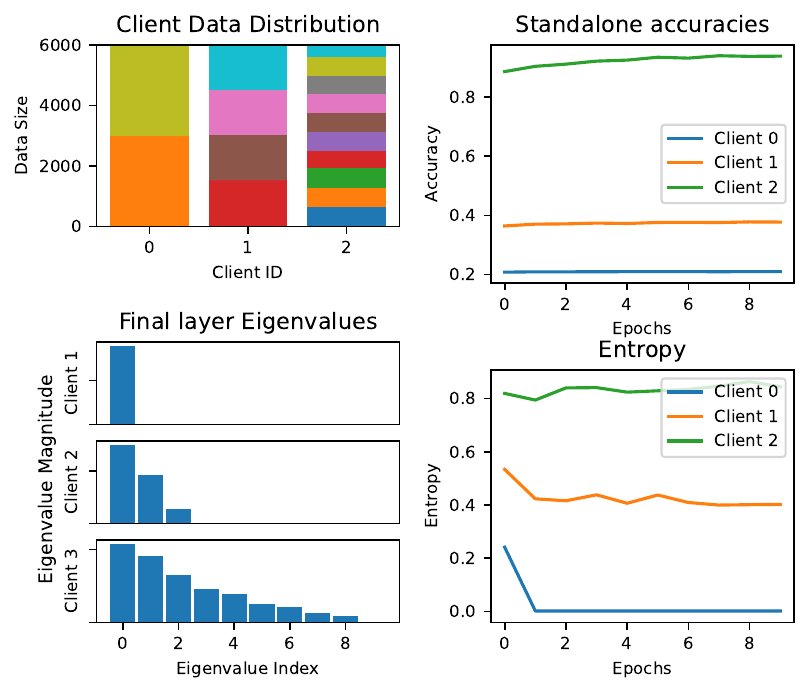}
    \caption{\textbf{Top left:} Three clients with equal amount of data but varying number of labels from the MNIST dataset. \textbf{Bottom left:} Eigenvalue spectrum corresponding to the final-layer updates of each client. \textbf{Top right:} Accuracies obtained by each client on a hold out test set as training progresses. \textbf{Bottom right:} Spectral entropy for each client as the training progresses.
    }
    \label{fig:entropy_motivation}
\end{figure}
\subsection{Spectral Entropy as a Data-Free Indicator}
\label{subsec:entropy}

The first component of our method is based on the von Neumann (spectral) entropy. We present a brief sketch of the relation between spectral entropy and its potential in assessing client utility.
Specifically, we turn to the phenomenon of \textit{Neural Collapse} presented by \cite{papyan_prevalence_2020}. In one of their propositions they show that the last-layer weights align with the class means and form a simplex ETF (See NC2 and NC3 in \cite{papyan_prevalence_2020}). The authors of \cite{zhu_geometric_2021} further prove that Simplex ETF corresponds to the global minima for unconstrained feature models. The authors of  \cite{markou2024guiding} utilize Neural Collapse during optimization for faster convergence. 

\noindent In our setting, the model update from client $i$ at round $t$ is given as $\Delta w_{i}^{(t)}$. We denote the weight matrix corresponding to the last layer, $L$, as the update matrix as $M_i^{(t)} = \Delta w_{i,L}^{(t)} $. Without loss of generality, we drop the superscript ${(t)}$ and subscript ${i}$ for further analysis.

\noindent For the update, $M\in\mathbb{R}^{C\times d}$, let $A := M M^\top \in \mathbb{S}_+^C$ be its class-space Gram matrix, where $C$ is the number of classes.
Let $\rho := A / \operatorname{Tr}(A)$ be the density matrix corresponding to $A$ with eigenvalues $\{p_j\}_{j=1}^C $, giving $\sum_j p_j = 1$ .
Then the von Neumann entropy of \(\rho\) is given as
\begin{equation}
S(\rho) \;=\; - \sum_{j=1}^C p_j \log p_j,
\end{equation}
Under the neural-collapse regime, the last-layer weight vectors align with class prototypes and span an almost-orthogonal frame of class directions $\{u_c\}_{c=1}^C$, \cite{papyan_prevalence_2020}. A client’s update decomposes along these directions, and the class–space energy allocated to direction $u_c$ is
\begin{equation}
\alpha_c \;:= \; u_c Au_c^\top = \; \big\| M^\top u_c \big\|_2^2 .
\end{equation}
\noindent Since $\{u_c\}_{c=1}^C$ are approximately orthogonal, $A \approx \sum_{c=1}^C \alpha_c \, u_c u_c^\top$
and the eigenvalue spectrum of $A$ is approximately $\{\alpha_c\}_{c=1}^C$.
Normalizing $\alpha_c $to sum to one, 
\begin{equation}
p_c \;=\; \frac{\alpha_c}{\sum_{k=1}^C \alpha_k},
\qquad
S(\rho) \;=\; -\sum_{c=1}^C p_c \log p_c,
\label{eq:entropy_proof}
\end{equation}
Therefore, the von Neumann entropy is approximately the Shannon entropy of the normalized class-space energies. It increases when the update’s energy spreads more evenly across class directions and decreases when it concentrates in a few directions. We demonstrate this phenomenon empirically in Fig.~\ref{fig:entropy_motivation} by computing the entropy scores for clients with the same amount of data but a varying number of class labels. The entropy scores correlate positively with the standalone accuracies of the clients.

\noindent In implementation, we use the WeightWatcher tool  \cite{martinPredictingTrendsQuality2021}, which internally computes entropy as:
\begin{equation}
    S_{i}^{(t)} = S\bigl(\tilde A_i^{(t)}\bigr) = -\sum_{j}\lambda_j \log \lambda_j. 
    \label{eq:entropy}
\end{equation}
where $\tilde A_i^{(t)} = \frac{A_i^{(t)}}{\|A_i^{(t)}\|_\mathrm{F}}$; This yields entropy values equivalent to those in Eqn. \ref{eq:entropy_proof} up to a scaling factor.

\noindent \emph{Utility implication.} In federated aggregation, clients  that possess a wider range of classes provide updates cover multiple class directions tend to reduce global loss more broadly and less redundantly. Neural Collapse theory shows that late-stage representations and classifier weights organize into near-orthogonal class directions. Because a client’s last-layer update decomposes along these directions, the von Neumann entropy measures how broadly its update energizes the class directions, providing a data-free proxy for utility in aggregation.


\noindent We construct our first algorithm, \textsc{SpectralFed}, first applying a momentum term to the estimates and then normalizing these smoothed entropy estimates to obtain the final client weights $c_i^{(t)}$. The complete \textsc{SpectralFed} method is described in Algorithm \ref{alg:spfd}. 
\begin{algorithm}[t]
    \caption{\texttt{SpectralFed}}\label{alg:spfd} 
    \renewcommand{\algorithmicrequire}{\textbf{Input:}}
    \renewcommand{\algorithmicensure}{\textbf{Procedure}}
    \begin{algorithmic}[1]
        \Require global weights: $w^0_{g}$, local learning rate $\eta$, no. of communication rounds $T$, no. of local epochs $E$, momentum factor: $\mu$, set of clients: $\mathcal{K} = \{1,2 \ldots n\}$ 
        \State Initialize client weights : $c^0_i \gets {1}/{n}, \forall i \in \mathcal{K}$ 
        \For{$t = 1 \ldots T$} 
            \State Broadcast global model:  ${w}^{(t)}_{i} \gets w^{(t)}_{g}  $ 
        
            \For {all clients $i \in \mathcal{K}$ in parallel}
                \For{$e = 1 \ldots E$}
                    \State Sample $z_{i}  \sim \mathcal{D}_i$ 
                    \State Compute gradient: $\nabla F_i \big(w^{(t)}_{i, e}, z_{i}\big)$ 
                    \State $w^{(t)}_{i, e+1} \gets $ SGD ($w^{(t)}_{i, e}, \nabla F_i, \eta$) 
                \EndFor 
            \EndFor
            \State Compute $S^{(t)}_{i}$ using Equations \ref{eq:entropy}, 
            \State $ \tilde{S}^{(1)} = S^{(1)} $ \quad $\tilde{S}^{(t)} = \mu \tilde{S}^{(t-1)}+ (1-\mu) S^{(t)}$ 
            \State $ c^{(t)}_{i}=  \tilde{S}^{(t)}_{i} / \sum_{i=1}^n \tilde{S}^{(t)}_{i}  $ 
            \State Update global model $w^{(t+1)}_{g} \gets \sum_{i=1}^n c^{(t)}_i w^{(t)}_{i, E}$ 
        \EndFor
    \end{algorithmic}
\end{algorithm}

\subsection{Complementary Class-specific Shapley Values}
\label{subsec:cssv}
While spectral entropy captures the spread of the updates well, it is less discriminative under scenarios where the class distribution is balanced but sample quantities differ as we show in section \ref{sec:results}. To complement the entropy score we leverage \emph{Class-specific Shapley values} (CSSV) proposed by the authors of ShapFed \cite{tastan2024redefining}.
Let $\hat{m}_{i,j}$ , $\hat{m}_{s,j}$ denote the $j$-th column
vectors of the final-layer update matrices, $M_{i}$,  $M_{g}$, for $j \in \{1,\dots C\}$. 
The per class contribution (CSSV) $\hat{\Gamma}_{i,j}$ of participant $i$ is the cosine similarity between $\hat{m}_{i,j}$ and $\hat{m}_{s,j}$ which is aggregated over $C$ classes to obtain a single scalar $\Gamma_{i}^{(t)}$ per client :
\begin{equation}
    \hat{\Gamma}_{i,j} =
    \cos \bigl(\hat{m}_{i,j},\hat{m}_{g,j}\bigr)
\quad; \quad
\Gamma_{i}^{(t)} = \frac{1}{C}\sum_{j=1}^{C}\hat{\Gamma}_{i,j}^{(t)}
\label{eq:cssv}
\end{equation}

\subsection{Rank-Adaptive Kalman Filtering (RAKF) of Contribution Metrics}
\label{sec:rankkalman}
In each communication round $t$, we obtain for every client $i$ two complementary indicators of contribution quality:
\begin{itemize}[leftmargin=*,topsep=1pt]
    \item \textbf{Final-layer gradient entropy} $S_{i}^{(t)}$, which measures the dispersion of the client’s final-layer update distribution.
    \item \textbf{Class-specific Shapley alignment} $\Gamma_{i}^{(t)}$, computed as the directional alignment between the client’s final-layer update and the global update.
\end{itemize}

\noindent Since the reliability of these metrics varies with client heterogeneity, we
introduce a \emph{Rank-Adaptive Kalman Filter} to produce a fused
contribution score $\hat{x}_{i}^{(t)}$ that automatically emphasizes the more
informative metric at each round.
We adopt the filter per client whose latent state $x_{i}^{(t)}$ models the “true’’ (but unobserved) contribution. Following the standard notations from section \ref{sec:prelim},
The process model is
\begin{equation}
\label{eq:proc_model}
x_{i}^{(t)} = x_{i}^{(t-1)} + w_{i}^{(t)}, 
\qquad w_{i}^{(t)}\sim \mathcal{N}(0,Q),
\end{equation}
where $Q$ is the process noise variance. For observations, we first standardize the smoothed entropy estimates ($\tilde{S}^{(t)}_{i}$) and smoothed CSSV ($\tilde{\Gamma}^{(t)}_{i}$) estimates to sum to one:
\begin{equation}
\label{eq:normalize}
    s^{(t)}_{i}=  \frac{\tilde{S}^{(t)}_{i}}{\sum_{i=1}^n \tilde{S}^{(t)}_{i}}; \quad 
     \gamma^{(t)}_{i}=  \frac{\tilde{\Gamma}^{(t)}_{i}}{ \sum_{i=1}^n \tilde{\Gamma}^{(t)}_{i}}
\end{equation}
At each round, we form the measurement vector 
\begin{equation}
\label{eq:meas_model}
y_{i}^{(t)} =
\begin{bmatrix}
s_{i}^{(t)} \\[3pt]
\gamma_{i}^{(t)}
\end{bmatrix}
= Hx_{i}^{(t)} + v_{i}^{(t)},
\quad v_{i}^{(t)}\sim\mathcal{N}(0,R_{i}^{(t)}), 
\end{equation}
where $H=[1\;1]^{\top}$.
To make the filter \emph{rank-adaptive}, we first compute the Spearman rank correlation between the predicted client estimates before the measurement update, $\mathbf{{\hat{x}}}^{(t)}_{|t-1} = [\hat{x}_{i|t-1}^{(t)}]^T$, and each of the incoming observations $\mathbf{{s}}^{(t)} = [s_{i}^{(t)}]^T$ and $\boldsymbol{{\gamma}}^{(t)} = [\gamma_{i}^{(t)}]^T$. 
We then map the observation noise covariance to inversely vary with these correlations, with $\epsilon$ as a small stability factor:
\begin{equation}
\label{eq:rupdate}
R^{(t)}=\operatorname{diag} \bigl( (1 - \rho_s^{(t)})+\epsilon,\;
(1 - \rho_\gamma^{(t)})+\epsilon \bigr) \forall i \in \mathcal{K}
\end{equation}
\noindent The filter schematic is illustrated in Figure \ref{fig:kalman} and the algorithm is outlined in Appendix A.
Intuitively, we seek to penalize observations that frequently change the ordering of the clients' contribution scores. If, say, the client rankings obtained from the entropy estimates varies more frequently than the CSSV estimates, it is declared less reliable and should contribute less towards the final clients' contribution estimates.
\begin{figure}
    \centering
    \includegraphics[width=\linewidth]{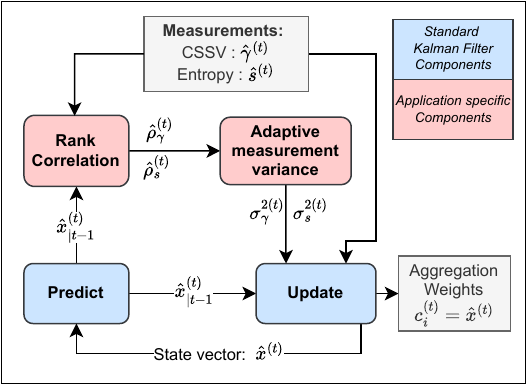}
    \caption{Schematic of the Rank Adaptive Kalman Filter used to fuse the CSSV estimates with the entropy estimates to get the final aggregation weights}
    \label{fig:kalman}
\end{figure}
\subsection{Putting It All Together: \texttt{SpectralFuse}} 
Our complete pipeline is illustrated in Fig.~\ref{fig:main}. Each client update is first processed to compute the spectral entropy $S_{i}^{(t)}$ and the Class-specific Shapley value $\Gamma_{i}^{(t)}$. The \textsc{SpectralFed} variant uses only the entropy to set aggregation weights, which is simple and effective in label-skew scenarios. The \textsc{SpectralFuse} variant feeds both $S_{i}^{(t)}$ and $\Gamma_{i}^{(t)}$ into the rank-adaptive Kalman Filter to produce a smoothed and reliable contribution estimate. This estimate is then used to weight client updates in the global aggregation step.
Our methodology is entirely server-side and data-free. It does not require access to any validation data, nor to self-reported client statistics, and it is robust to heterogeneous client behavior. 



 

\begin{algorithm}[t]
    \caption{\texttt{SpectralFuse}}
    \renewcommand{\algorithmicrequire}{\textbf{Input:}}
    \renewcommand{\algorithmicensure}{\textbf{Procedure}}
    \begin{algorithmic}[1]
        \Require  $w_{g,0}$, $\eta$, $T$,  $E$, $\mu$, $\mathcal{K} = \{1,2 \ldots n\}$ 
        \State Initialize client weights : $c^0_i \gets {1}/{n}, \forall i \in \mathcal{K} $ 
        \State Initialize \texttt{filter} $\gets$ \texttt{Rank Adaptive Kalman Filter()} 
        \For{$t = 1 \ldots T$} 
            \State Broadcast global model:  ${w}^{(t)}_{i} \gets w^{(t)}_{g} \forall i \in \mathcal{K}$ 
            \State $w^{(t)}_{i, E} , \hat{S}^{(t)}_{i} \gets $ Run Steps 4-12 of \texttt{SpectralFed} \ref{alg:spfd}

            \State Compute $ s^{(t)}_{i}, \gamma^{(t)}_{i}$ from Equation \ref{eq:normalize}
            \State $\hat{x}^{(t)}_{i} \gets$ \texttt{filter}($\gamma^{(t)}_i, s^{(t)}_i$)
            \State Normalize: $ c^{(t)}_{i}=  \hat{x}^{(t)}_{i} / \sum_{i=1}^n \hat{x}^{(t)}_{i} $ 
            \State Update global model $w^{(t+1)}_{g} \gets \sum_{i=1}^n c^{(t)}_i w^{(t)}_{i, E}$ 
        \EndFor
    \end{algorithmic}
\end{algorithm}

\section{Implementation Details}
\label{sec:implementation}
\begin{table*}[t!]
\centering
\caption{Average Pearson Correlation of client weights $c_i$ with standalone final accuracies of methods across datasets and splits. The best results are shown in \textbf{bold} and the second best results are \underline{underlined}.\label{tab:correlation}}
\begin{tabular}{l l c c c c c c}
 
\toprule
\textbf{Dataset} & \textbf{Split}  & \textbf{FedAvg} & \textbf{CGSV} & \textbf{ShapFed} & \textbf{SpectralFed} & \textbf{SpectralFuse} \\
\midrule

\multirow{5}{*}{CIFAR-10} 
& Only Label Skew & -0.01 ± 0.02 & 0.88 ± 0.03 & 0.89 ± 0.02 &\underline{ 0.95 ± 0.02} & \textbf{0.97 ± 0.01} \\
& Step Label Skew & -0.01 ± 0.02 & 0.95 ± 0.02 & 0.93 ± 0.05 & \underline{0.96 ± 0.01} & \textbf{0.97 ± 0.01} \\
& Step Quantity   &  0.03 ± 0.04 & \underline{0.92 ± 0.06} & 0.91 ± 0.03 & 0.81 ± 0.02 & \textbf{0.95 ± 0.01} \\
& Dirichlet ($\alpha$ = 0.1  ) & -0.01 ± 0.02 & \textbf{0.74 ± 0.19} & 0.58 ± 0.53 & 0.72 ± 0.20 & \textbf{0.80 ± 0.14} \\
& Dirichlet ($\alpha$ = 0.01)  & -0.00 ± 0.02 & 0.88 ± 0.04 & 0.94 ± 0.03 & \textbf{0.97 ± 0.03} &\textbf{ 0.97 ± 0.03} \\ [0.5em]
\midrule
\multirow{5}{*}{CIFAR-100}
& Only Label Skew & 0.02 ± 0.01 & 0.38 ± 0.33 & 0.81 ± 0.13 & \textbf{0.90 ± 0.14} & \underline{0.88 ± 0.15 }\\
& Step Label Skew & 0.02 ± 0.02 & 0.56 ± 0.11 & 0.87 ± 0.08 & \underline{0.90 ± 0.04} &\textbf{ 0.94 ± 0.02} \\
& Step Quantity & -0.01 ± 0.02 & 0.15 ± 0.17 & 0.92 ± 0.03 & \underline{0.92 ± 0.02} & \textbf{0.95 ± 0.01} \\
& Dirichlet ($\alpha$ = 0.1  ) & 0.02 ± 0.03 & -0.07 ± 0.24 & 0.08 ± 0.61 & \underline{0.68 ± 0.15} & \textbf{0.74 ± 0.05} \\
& Dirichlet ($\alpha$ = 0.01) & -0.02 ± 0.06 & 0.51 ± 0.03 & 0.79 ± 0.22 & \underline{0.94 ± 0.01} & \textbf{0.95 ± 0.03} \\ [0.5em]
\midrule
FEMNIST & Natural split  & -0.00 ± 0.01 & 0.12 ± 0.03 & \textbf{0.41 ± 0.03} & 0.35 ± 0.02 & \underline{0.40 ± 0.01}  \\
FedISIC & Natural split  & -0.01 ± 0.01 &\underline{0.88 ± 0.01} &\textbf{ 0.92 ± 0.01} & 0.82 ± 0.01 & 0.86 ± 0.01 \\
\bottomrule
\end{tabular}
\end{table*}

\noindent\textbf{Datasets and Models.}
For CIFAR-10, we use a randomly initialized 5-layer CNN model , while for the CIFAR-100 experiments we use a randomly initialized \text{ResNet-50} model. For FEMNIST we use a 4-layer MLP and for the FedISIC dataset, we used the ImageNet-1k pretrained \text{ViT-B/16} model. 

\noindent\textbf{Hyperparameters.}
We simulated $5$ clients for CIFAR-10/100 and $6$ clients for FedISIC. We subsample 100 clients for the FEMNIST dataset to prevent memory overruns.
Each client performed one local epoch per communication round. 
The CIFAR experiments were run for $200$ rounds, FEMNIST  for $100$ rounds, and FedISIC for $300$ rounds. 
Unless otherwise stated, we used stochastic gradient descent without momentum, initial learning rate $0.1$ for CIFAR-10/100 and FEMNIST and $0.001$ for FedISIC, and cosine annealing scheduler with a final learning rate of $10^{-6}$. The loss function used was cross-entropy loss.
We use a batch size of $64$ for all settings. For the Kalman Filter, we set the process noise variance $Q=10^{-4}$ and the measurement noise stability factor $\epsilon=10^{-3}$. An ablation study on the sensitivity of $Q$ and $\epsilon$ is shown in Appendix. 
The momentum factor for smoothing the entropy and CSSV estimates was set to $\mu=0.9$ similar to \cite{tastan2024redefining}.

\noindent\textbf{Data Splits.}
We considered five types of non-IID splits for CIFAR-10 and CIFAR-100 along with the standard IID split:
(i) Only Label Skew: imbalanced labels with equal data points,
(ii) Step Quantity Skew: progressively increasing data quantity with balanced labels,
(iii) Step Label Skew: progressively increasing label variety and data quantity,
Dirichlet splits with (iv) $\alpha=0.1$,
and (v) $\alpha=0.01$.
The FEMNIST and FedISIC datasets use the naturally provided client split.

\noindent\textbf{Entropy and CSSV Computation.}
In each communication round, we compute the von Neumann entropy $S_i$ of the spectral distribution of the final-layer gradient update for each client $i$. This is done using the WeightWatcher tool \cite{martinPredictingTrendsQuality2021}. 
We also compute the class-specific Shapley value vector $\Gamma_i$ using the approach of \cite{tastan2024redefining}. 

\noindent\textbf{Baselines}.
We compare our methods against the following baselines: FedAvg \cite{mcmahanCommunicationEfficientLearningDeep2017}, CGSV \cite{xuGradientDrivenRewards2021}, and ShapFed \cite{tastan2024redefining}. 
For FedAvg, we consider the variant with uniform weighting (FedAvg-uniform) for a fair comparison with data-free approaches and no self-reported metadata. Both CGSV and ShapFed are data-free methods that do not require auxiliary validation data, but provide client rewards commensurate to their contribution estimates. Since we focus mainly on the contribution estimates, we disable the client reward mechanisms in CGSV and ShapFed for a fair comparison. 

\noindent\textbf{Evaluation Protocol.}
We mainly report the Pearson correlation between the client weights $c_i$ and the standalone accuracies of each client. The standalone accuracies are obtained when each client independently trains on its own data and evaluates on a held-out test set. We also report the global model accuracy obtained at the end of training for each method and each split.
All reported metrics are reported as mean and standard deviation over 3 seeds for CIFAR100, FEMNIST, and FedISIC and 5 for CIFAR10.

\section{Results}
\label{sec:results}

The results of our experiments are summarized in Tables~\ref{tab:correlation} and~\ref{tab:main_results}. Table~\ref{tab:correlation} examines how well each contribution estimation method correlates with standalone client accuracy. CGSV and ShapFed attain strong correlations in several settings but are inconsistent: CGSV performs well on quantity-skew settings on CIFAR-10 yet collapses under more heterogeneous CIFAR-100 splits, while ShapFed performs relatively better under label skew settings. In contrast, our entropy-based methods (SpectralFed and SpectralFuse) provide consistently high correlations. On CIFAR-10, SpectralFuse remains above 0.95 across nearly all non-IID splits, with SpectralFed close behind. A similar pattern holds on CIFAR-100, where the entropy-based methods clearly outperform the baselines in the heterogeneous regimes, maintaining substantial positive correlation. Even on the naturally partitioned FEMNIST and FedISIC datasets, SpectralFed and SpectralFuse remain competitive, indicating that the entropy signal transfers to real-world non-IID distributions. Furthermore, RAKF allows SpectralFuse to maintain sustained high correlations across all settings. To illustrate the consistency of the entropy-derived weights, we show the per-round Pearson correlation for the proposed methods and baselines in Figure \ref{fig:corr_analysis}. Further analysis on the operating principle of the filter is provided in the Appendix.

\begin{table*}[t!]

\centering
\caption{Predictive performance of the methods across datasets and splits. * CGSV does not compute the global model, so we report the average client performance. \textdagger We report the balanced accuracy for FedISIC to account for the label imbalance in its test set. The best results are shown in \textbf{bold} and the second best results are \underline{underlined}.\label{tab:main_results}}
\begin{tabular}{l l c c c c c}
 
\toprule

{\textbf{Dataset} }& {\textbf{Split}} &{ \textbf{FedAvg}} & {\textbf{CGSV*}} & {\textbf{ShapFed}} & \textbf{SpectralFed} & \textbf{SpectralFuse} \\
\midrule

\multirow{6}{*}{CIFAR-10} 

& IID & \textbf{82.45 ± 0.11} & 75.48 ± 0.27 & 81.99 ± 0.19 & \underline{82.27 ± 0.09} & 82.25 ± 0.18 \\
& Only Label Skew & 67.97 ± 1.71 & 36.72 ± 0.81 & 67.81 ± 0.79 & \underline{69.07 ± 1.42} & \textbf{69.43 ± 0.84} \\
& Step Label Skew & 81.95 ± 0.24 & 55.75 ± 1.06 & 81.73 ± 0.34 & \textbf{82.35 ± 0.28} & \underline{82.15 ± 0.23} \\
& Step Quantity & 82.66 ± 0.23 & 75.16 ± 0.34 & 82.12 ± 0.29 & \textbf{82.81 ± 0.20} & \underline{82.75 ± 0.15} \\
& Dirichlet ($\alpha$ = 0.1  ) & \textbf{76.42 ± 1.26} & 36.49 ± 2.16 & 75.72 ± 1.17 & 75.76 ± 1.13 & \underline{76.24 ± 0.89} \\
& Dirichlet ($\alpha$ = 0.01)  & \textbf{67.14 ± 2.20} & 21.73 ± 0.85 & 58.14 ± 3.22 & 65.21 ± 2.35 & \underline{66.18 ± 2.38} \\ [0.5em]

\midrule
\multirow{6}{*}{CIFAR-100}
& IID & \textbf{43.19 ± 1.22} & 25.69 ± 0.46 & 40.39 ± 0.26 & 41.57 ± 1.42 & \underline{42.70 ± 2.38} \\
& Only Label Skew & \underline{34.67 ± 0.38} & 15.66 ± 0.15 & 35.78 ± 0.45 & \textbf{35.62 ± 0.84} & 34.28 ± 1.63 \\
& Step Label Skew &\textbf{ 43.90 ± 0.27} & 21.53 ± 0.48 & 41.65 ± 0.55 & \underline{43.88 ± 0.13} & 43.63 ± 1.03 \\
& Step Quantity & \underline{42.66 ± 1.11} & 25.44 ± 0.43 & 41.83 ± 1.15 &\textbf{ 42.97 ± 0.37} & 42.69 ± 3.16 \\
& Dirichlet ($\alpha$ = 0.1  ) & 39.01 ± 0.45 & 16.04 ± 0.36 & \underline{39.42 ± 1.06} & 39.20 ± 1.33 & \textbf{40.45 ± 1.71} \\
& Dirichlet ($\alpha$ = 0.01)  & 33.57 ± 1.73 & 11.89 ± 0.90 & \textbf{36.33 ± 0.16} & \underline{35.99 ± 0.59} & 34.87 ± 0.46\\[0.5em]
\midrule
FEMNIST & Natural split  & 60.64 ± 0.26 & 29.98 ± 0.22 & 60.82 ± 0.22 & \textbf{61.17 ± 0.27} & \underline{61.06 ± 0.23} \\

FedISIC\textdagger & Natural split & 51.26 ± 0.98 & 37.23 ± 0.22 & 51.58 ± 1.92 & \textbf{55.59 ± 0.67} &\underline{ 55.34 ± 0.48} \\
\bottomrule
\end{tabular}
\end{table*}

\begin{figure}
    \centering
    \includegraphics[width=\linewidth]{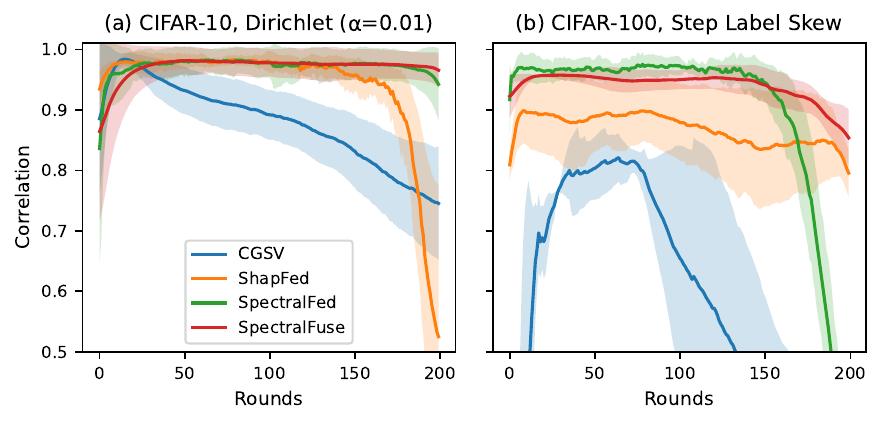}

    \caption{SpectralFed (green) shows consistent Pearson correlation with the client standalone accuracies across rounds. SpectralFuse (red) benefits from fusing both CSSV and Entropy signals and improves correlation by resisting frequent rank changes. }
    \label{fig:corr_analysis}
\end{figure}
\noindent Table~\ref{tab:main_results} compares the predictive performance of the global model under each aggregation scheme. In the IID condition, all methods perform reasonably well. Under non-IID splits, however, the advantage of entropy-based weighting is more evident. On CIFAR-10, SpectralFed and SpectralFuse perform well in the Only Label Skew setting, while maintaining parity in Step Quantity and Step Label Skew. 
Under the more challenging Dirichlet partitions, entropy weighting remains comparable to or slightly better than the baselines. 
A similar trend appears on CIFAR-100. On FEMNIST and FedISIC, the benefits are more pronounced. SpectralFed achieves a balanced test accuracy of 55.59\%, exceeding FedAvg by over four points and outperforming ShapFed. 
These findings indicate that the data-free entropy signal not only aligns with client quality but can also translate into tangible improvements in the aggregated model's accuracy. 
However, the absolute gains are often modest, suggesting that while finding good contribution estimates may help with fairness and incentives distribution, it may not necessarily translate to superior global model performance. We discuss this further in the next section. 

\noindent 
Across CIFAR-10/100, FEMNIST, and FedISIC, entropy-derived weights strongly correlate with standalone client performance and yield improvements in global accuracy over uniform weighting. 
The Kalman Filter fuses the signals from CSSV and entropy to provide a smooth and noise robust estimate. 
The modular nature of the filter allows its extension to potentially work with multiple input signals.
The evidence supports that von Neumann entropy in conjunction with alignment-based methods, can serve as a reliable proxy for client contributions. 
\section{Discussion}
\label{sec:discussion}

\noindent\textbf{Why does spectral entropy work?}
The final layer encodes the task's class geometry and aligns with the mean class directions under the neural-collapse regime. Clients whose data are informative for decision boundaries tend to induce weight updates with richer spectral profiles. 
This helps in non-IID regimes where valuable clients may contribute updates that deviate from the cohort average (e.g., minority labels or rare features). Unlike similarity based approaches that reward conformity to the mean gradient, the entropy view can value these informative, diverse updates without access to any data or labels.  The strong empirical correlation with standalone accuracy supports this interpretation. On the other hand, similarity/alignment based approaches benefit in scenarios with diverse client quantities, as the best clients produce bigger updates and dominate the aggregate gradient. However, under label-diversity, they may struggle to account for the full range of client contributions.

\noindent\textbf{Interpreting the accuracy gains.}
The accuracy improvements we observe are typically modest relative to FedAvg on CIFAR-10/100, and a bit larger on FEMNIST and FedISIC. We hypothesize that two factors help explain the modest gains: (i) since we supply the full global model to all the clients, all clients eventually train on the same model and any gains obtained from commensurate weight assignment eventually taper off as the training ends. (ii) Our method adjusts aggregation weights but leaves local training unchanged, which may be the primary bottleneck for the global model performance. We suspect that the combination of these factors leads to the observed improvements. Further research on these hypotheses may help validate them.

\noindent\textbf{Future directions.}
A practical benefit of data-free contribution score is that it can drive \emph{incentive mechanisms}  without auxiliary validation sets.
Enhanced aggregation schemes that can better utilize the contribution estimates to improve global model performance are a potential avenue for future exploration. 
\noindent In summary, entropy-based contribution estimation provides a simple, privacy-preserving, and effective signal for weighting clients in federated learning. Combined with alignment-based approaches using filtering, it can help ensure that client contributions are accurately reflected in the training process.

\section*{Acknowledgements}
This material is partly based on work supported by the Office
of Naval Research N00014-24-1-2168.

{
    \small
    \bibliographystyle{ieeenat_fullname}
    \bibliography{library}
}

\clearpage
\setcounter{page}{1}
\maketitlesupplementary

\appendix 

\section{Rank Adaptive Kalman Filter Algorithm}
\label{app:rkhf}

\begin{algorithm}[H] 
    \caption{\texttt{Rank Adaptive Kalman Filter}} \label{alg:kf}
    \renewcommand{\algorithmicrequire}{\textbf{Initialize:}}
    \renewcommand{\algorithmicensure}{\textbf{Input:}}
    \begin{algorithmic}[1]
        \Require $x^0_{i}, P^0, Q, \epsilon$ 
        \Ensure $\hat{x}^{(t-1)}_{i}, s^{(t)}_{i}, \gamma^{(t)}_{i} $
        \State \textbf{Predict:}  $\hat{x}^{(t)}_{i|t-1}$ using Equation \ref{eq:kf_pred}
        \State \textbf{Update}:
        \State $\rho^{(t)}_{s} = $ \texttt{Spearman} ($\mathbf{{\hat{x}}}^{(t)}_{|t-1}, \mathbf{s}^{(t)}$)
        \State ${\rho}^{(t)}_{\gamma} = $ \texttt{Spearman} ($\mathbf{\hat{x}}^{(t)}_{|t-1}, \boldsymbol{\gamma}^{(t)}$)
        \State $y^{(t)}_{i} = (s^{(t)}_{i}, \gamma^{(t)}_{i})^{\top}$
        \State Update $R_i^{(t)} $ from Equation \ref{eq:rupdate}
        \State Compute: $\hat{x}^{(t)}_{i}$ using Equation \ref{eq:kf_upd} \\
        \Return $\hat{x}^{(t)}_{i}$
    \end{algorithmic}
\end{algorithm}

\noindent
We provide an extended analysis of the RAKF proposed in section \ref{sec:rankkalman} here. The filter fuses two complementary, data-free signals :(i) the spectral (von Neumann) entropy of the final-layer update and (ii) the class-specific Shapley alignment (CSSV), into a single per-client contribution estimate that is stable over rounds yet responsive to persistent changes. Algorithm \ref{alg:kf} complements Figure \ref{fig:kalman} to illustrate the filter methodology. 

\begin{figure}[!h]
    \centering

    \includegraphics[width=\linewidth]{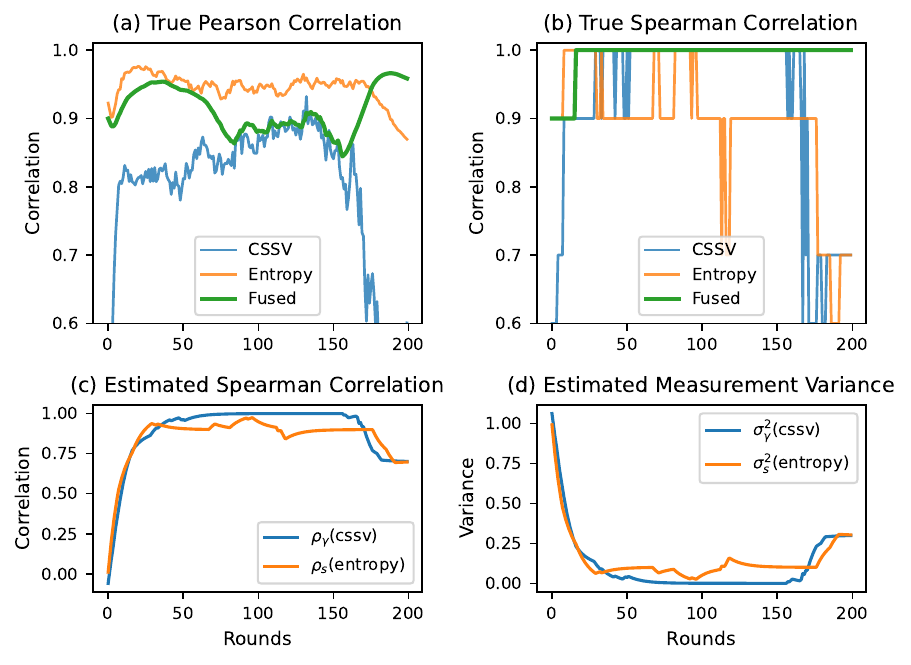}

    \caption{The RAKF's working principle demonstrated for a run on the Dirichlet ($\alpha$ = 0.1) split of CIFAR-10. (a) The RAKF fused weights smoothen and track the true Pearson correlation trend of the perceived reliable signal. (see ‘Fused’ vs. ‘CSSV’) (b) Fused weights resist rank changes and therefore maintain a stable Spearman Rank correlation profile over rounds. (c) The filter estimates the rank correlation between its previous state and each incoming signal. (d) The signal that has a higher estimated rank correlation has lower variance and is considered more trustworthy.}
    \label{fig:filter_analysis}
\vspace{-0.5em}
    
\end{figure}

\medskip
\noindent
We demonstrate the filter operation process in Figure \ref{fig:filter_analysis} by analyzing the internal values of the filter.
In Figure \ref{fig:filter_analysis} (a), the filter initially tracks the correlation trajectory of the entropy and later transitions to CSSV.
Figures \ref{fig:filter_analysis} (b-c) illustrate this: we observe that while the entropy-based rankings are initially consistent, they exhibit greater variation between rounds 50-150.
This, in turn, reduces its variance (Figure \ref{fig:filter_analysis} (d)), and therefore the fused weights are more dependent on the CSSV signal.
An key caveat is that the filter does not have access to ground-truth correlations, as the client utility (standalone accuracy) is unknown to the server.
Therefore, it may trust a signal as long as it is consistent, even if it may have a lower actual correlation. This is evident in Figure \ref{fig:filter_analysis} (a), where the signal with lower correlation (CSSV) is more trusted due to its consistency despite having a lower true correlation score.
Together, these plots illustrate the intended behavior of the filter, which initially reduces rank volatility but adapts to the signals when a sustained regime change occurs.


\section{Filter hyperparameter sensitivity}
\label{app:hparam}
\begin{figure}[!h]
    \centering
    \includegraphics[width=\linewidth]{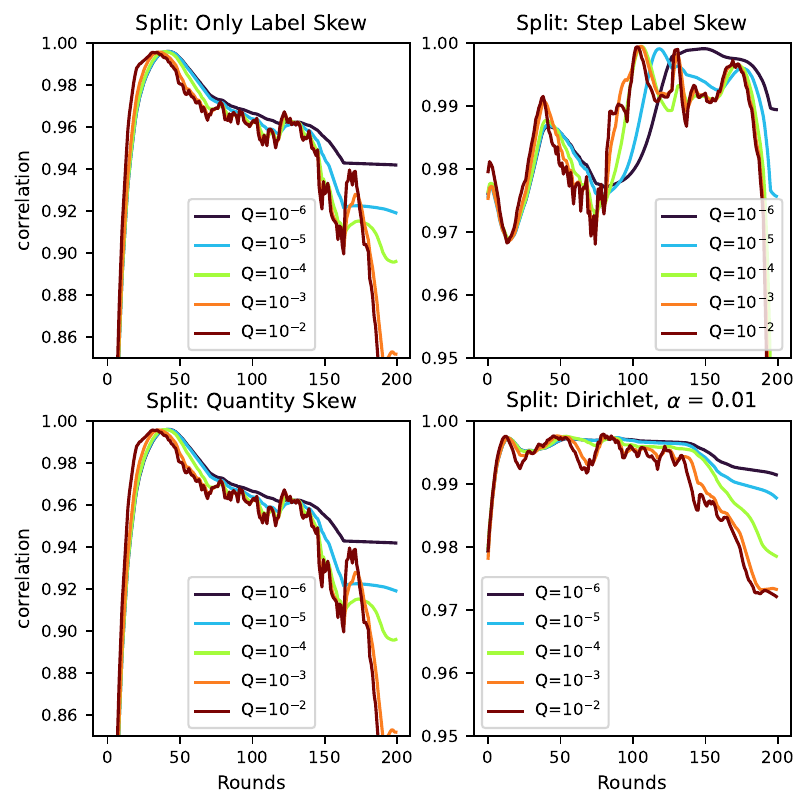}

    \caption{Effect of choice of filter process noise covariance $Q$ on the per-round Pearson correlation with standalone accuracies for different splits of the CIFAR-10 dataset. Lower values of $Q$  (see $Q=10^{-6}$) yield smoother but stagnant curves as the filter `trusts' its internal process model of persistent client ranks more.}
    \label{fig:q_abl}
\vspace{-0.5em}
\end{figure}

\begin{figure}[!h]
    \centering
    \includegraphics[width=\linewidth]{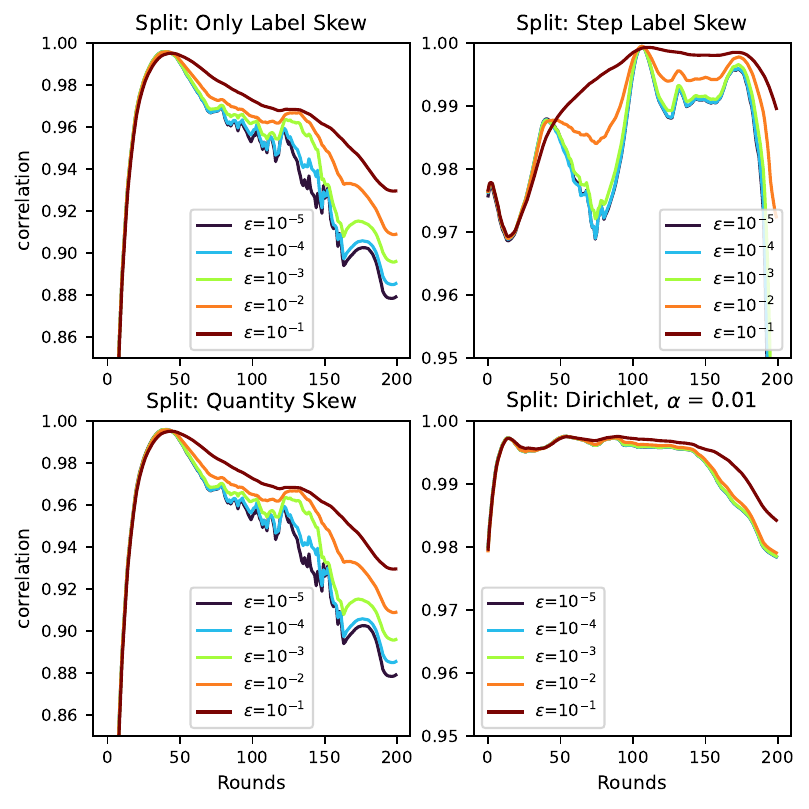}

    \caption{Effect of choice of measurement variance floor  $\epsilon$ on the per-round Pearson correlation with standalone accuracies for different splits of the CIFAR-10 dataset.  Higher values of $\epsilon$  raise the variance floor for the measurements (entropy and CSSV) and cap the filter's confidence in the measurement signals. The fused weights do not respond to rapid changes in the signals, resulting in a smoother correlation curve  (see $\epsilon=10^{-2}$)}
    \label{fig:ep_abl}
\end{figure}

\begin{figure}[!h]
    \centering
    \includegraphics[width=\linewidth]{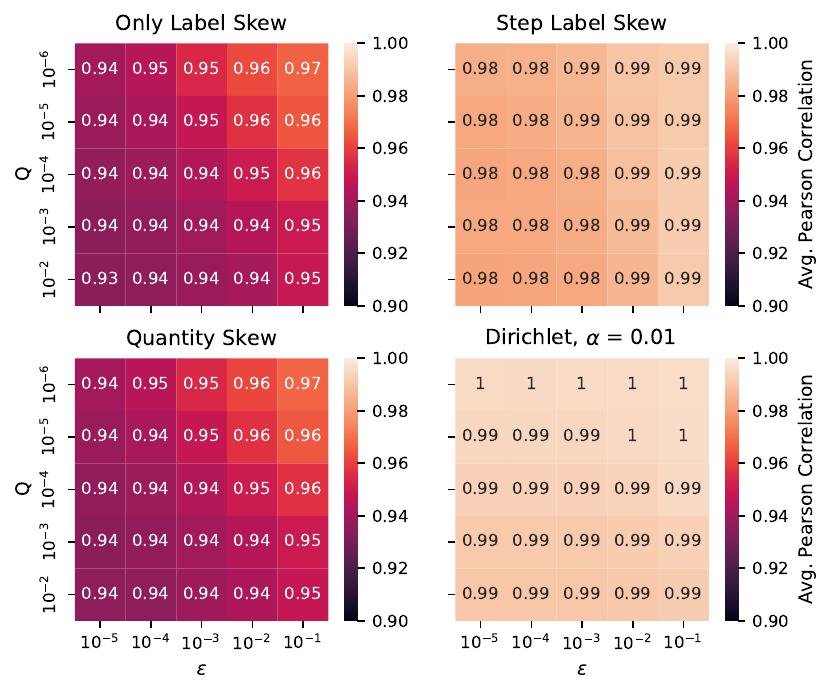}

    \caption{Sensitivity of choice of $Q$ and $\epsilon$ on the average Pearson correlation of the fused weights and standalone accuracies for different splits on the CIFAR-10 dataset. Average values of correlation are robust with respect to changes in filter hyperparameters. }
    \label{fig:sensitivity}
\vspace{-0.5em}   
\end{figure}

To assess the impact of filter hyperparameters $Q$ and $\epsilon$, on weight correlations with standalone accuracies, we provide a brief analysis of their sensitivity.

\noindent
\textbf{Process noise covariance $Q$.} In Figure \ref{fig:q_abl}, we tested the impact of varying the process noise covariance $Q$ on correlation performance over rounds. In order to interpret the results, we revisit the definition of $Q$ given in Equation \ref{eq:proc_model}. $Q$ is the estimate of the covariance of the normally distributed process noise in the state model. In a sense, it is a prior estimate of the fidelity of the process model that the filter is operating under. The better a process is modeled, the lower the value of $Q$ should be. In our case, the process model  is given as $x_{i}^{(t)} = x_{i}^{(t-1)} + w_{i}^{(t)}$ 
, where $x_{i}^{(t)}$ is the estimate of the weights of the client $i$ in round $t$. This implies that the client weights should remain static. In principle, this is a sound model as the clients' utility is expected to remain constant under the assumption that no client adds any additional data during the training process. However, given that the filter initializes under a uniform weight for each client, the client state should be allowed to change based on external heuristics. The choice of process noise covariance $Q$ governs how fast the state should respond to external signals when they differ from the internal state. We observe this phenomenon in practice in Figure \ref{fig:q_abl}. We observe that lower values of Q smooth the correlation curves and avoid the noise in the external signals (entropy and CSSV). Although this is a desirable behavior if the initial correlation is high (as in the Only Label Skew and Quantity Skew cases), it can be problematic when early correlations are poor and improve in later rounds (as in the Step Label Skew case). An intermediate value of $Q$ provides a good balance between these behaviors.
Across most splits, a broad middle band of values yields a similar performance, indicating low sensitivity in practice.

\medskip
\noindent
\textbf{Measurement variance floor $\epsilon$.}
Similarly to the study of the sensitivity of the process noise covariance, we study the impact of the choice of $\epsilon$. To recap, the measurement variance floor provides a lower bound to the measurement noise covariance matrix $R$ as described in Equation \ref{eq:rupdate}. The measurement noise covariance acts as a counterweight to the process noise covariance $Q$. In a typical Kalman Filter design, it provides a fixed prior on the covariance of external measurements. In the case of RAKF, $R^{(t)}$ is time-dependent and varies inversely with the rank correlation of the measurement signals (Equation \ref{eq:rupdate}). Intuitively, a higher value of $R$ implies a relatively lower confidence in the external measurements as compared to the process model. Higher $\epsilon$ results in a generally higher value of $R$. Figure \ref{fig:ep_abl} captures the effect of changing $\epsilon$ on the true Pearson correlation over rounds for different splits of the CIFAR-10 dataset. The empirical results validate this hypothesis,
as higher values of $\epsilon$ yield a delayed, more sluggish response to entropy/CSSV.


\medskip
\noindent
Although the choice of $Q$ and $\epsilon$ affects the behavior and performance of the RAKF, the variation observed in the correlation values when aggregated over rounds remains negligible. We capture this in Figure \ref{fig:sensitivity}, which shows the variation in the average correlation values as a function of  $(Q,\epsilon)$. Correlations remain high over wide regions of the grid for all splits, with very small variations for all splits. Practically, one can pick $Q$ to target desired smoothness and set a modest $\epsilon$ to avoid over-trusting rapidly fluctuating signals. We therefore show that the reported results are reasonably robust to the choice of filter hyperparameters.


\section{Free-rider detection}
\label{app:free-rider}
Contribution estimation schemes should not only be effective in identifying client utility under data heterogeneity but also should be able to detect scenarios where there are non-contributing clients or free-riders. A free-rider could be modeled in several ways. In the trivial scenario, it could be a client that does not perform training but participates only to obtain the global model or participation rewards. At the other extreme, a free rider could be malicious and attempt to actively derail the FL process. The first scenario can be easily identified by a simple client gradient norm check. The latter model requires further assumptions about a white-box, grey-box, or black-box knowledge setting and is better suited for treatment under the robust FL paradigm. 
We simulate a non-adversarial free-rider that participates by duplicating a very small amount of data to match the average cohort data size.
Any method that correlates with the amount of data only would not be able to distinguish this free-rider. 

\begin{figure}[!h]
    \centering
    \begin{subfigure}{\linewidth}
    \centering
    
    \includegraphics[width=\linewidth]{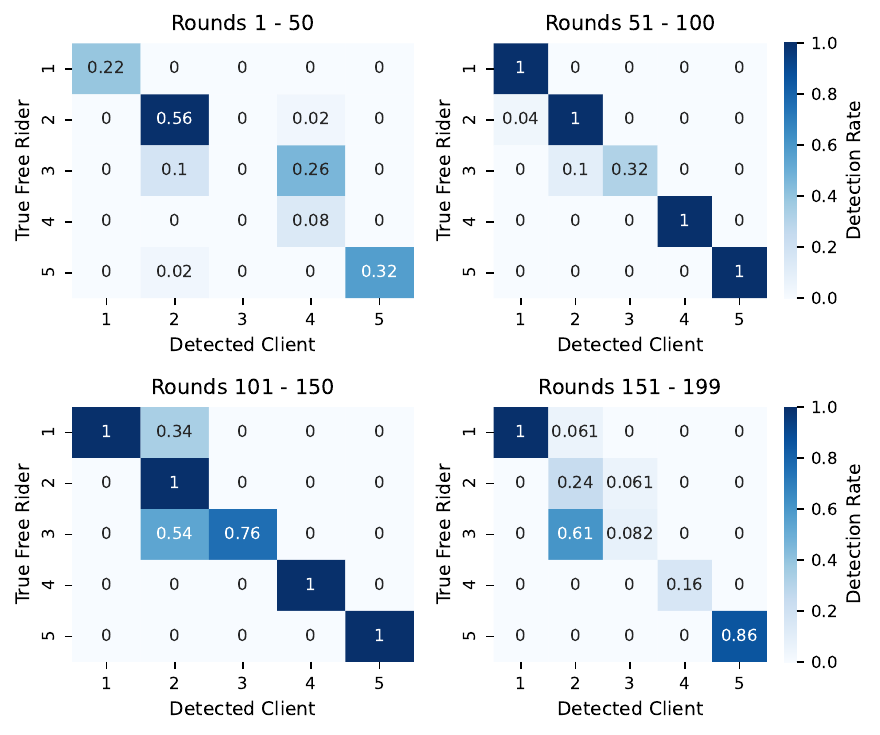}

    \caption{{SpectralFed} }
    \label{fig:free_rider_specfed}
    
    \end{subfigure}
    

    \centering
    \begin{subfigure}{\linewidth}
    \centering
    
    \includegraphics[width=\linewidth]{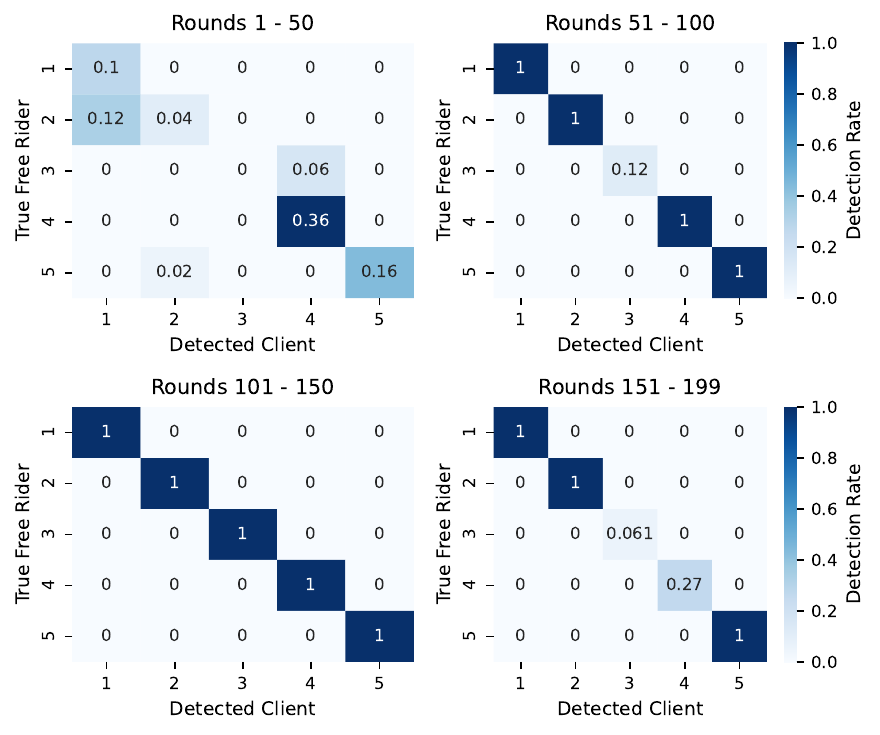}

    \caption{{SpectralFuse}}
    \label{fig:free_rider_specfuse}
    \end{subfigure}
    \caption{Free-rider detection accuracy of the proposed methods at four different intervals of training. Each row denotes an independent federated training procedure, with the y-axis indicating which client is the free rider. The values in the grid represent the rate of detecting the client, denoted on the x-axis as the free rider. High values on the diagonal indicate correct detections, while off-diagonal values indicate false positives. }
    \label{fig:free_rider}
\vspace{-0.5em}
\end{figure}

\medskip
\noindent In order to detect the free-rider from the contribution estimates, we first transform the estimates to the real space (contribution estimates are compositional as they have a fixed sum, one, and thus lie on a simplex). We then compute the robust Z-score $z_i = (x_i - \text{median}(x_i)/\text{MAD})$ where MAD is the Median Absolute Deviation. Free-riders are clients that are flagged as outliers in the Z-score. We run 5 experiments on the Dirichlet-partitioned ($\alpha$ = 1.0) CIFAR-10 dataset for both SpectralFed and SpectralFuse. One out of the 5 clients is the free rider in each experiment. We show the fraction of detections for each client in four 50-round intervals in Figure \ref{fig:free_rider}.


\noindent
For both SpectralFed (Fig. \ref{fig:free_rider_specfed}) and SpectralFuse (Fig. \ref{fig:free_rider_specfuse}), free-riders are detected with high accuracy towards the middle and later phases of training.
We observe lower early-phase detection with occasional off-diagonal false positives, likely because the weights are not yet well-separated.
These results indicate that simple thresholding of our data-free scores is effective for free-rider screening, without the need for auxiliary validation data or self-reported metadata.
\begin{figure*}[!h]
    \centering
    \includegraphics[width=\linewidth]{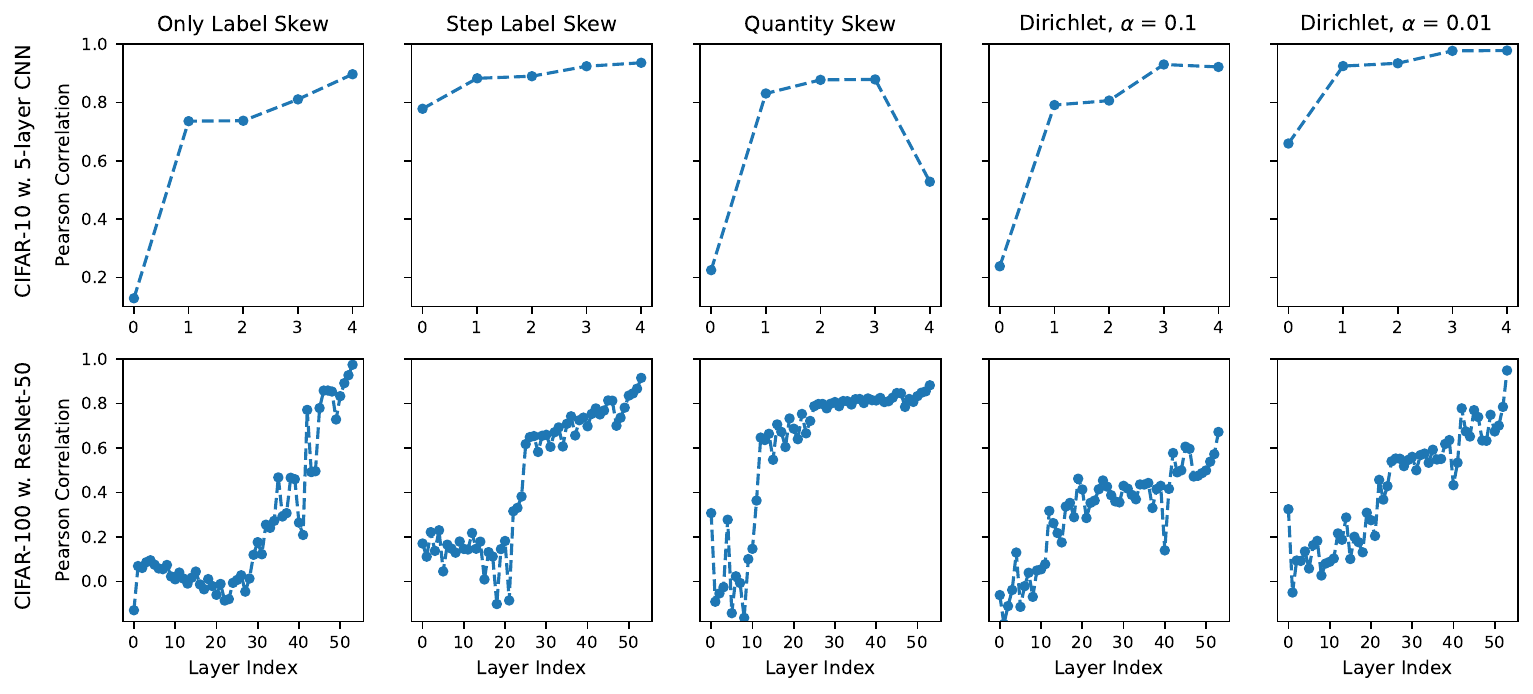}

    \caption{Average Pearson correlation of the entropy of the gradient for each layer with the standalone accuracy. The top row displays the average correlation for the 5-layer CNN used on the CIFAR-10 dataset, with each entry corresponding to a specific split. The x-axis denotes the layer index. The bottom row shows a similar analysis for the ResNet-50 model used on the CIFAR-100 dataset. We observe a general trend of increasing correlation as we move deeper in the network, with the final layer usually showing the highest correlation.}
    \label{fig:layer_wise_entropy}
\end{figure*}
\section{Choice of final layer for entropy evaluation}
\label{app:additional}

\noindent
In this section, we motivate the choice of using the final layer of the gradients for the entropy computation with empirical evidence. Figure \ref{fig:layer_wise_entropy} 
shows the average Pearson correlation between entropy and standalone accuracy as a function of layer depth for a 5-layer CNN on CIFAR-10 and ResNet-50 on CIFAR-100.
We observe a generally increasing correlation as we move deeper in the model with the highest correlations observed in the final layer. This provides strong empirical grounds for choosing the final layer only, rather than, e.g., averaging across layers
Furthermore, these results align with the theoretical basis for the role of the classifier head (final layer) in encoding the class geometry, as discussed in subsection \ref{subsec:entropy}. When a client’s data is uniform and diverse in labels, the final layer update exhibits a richer and more balanced spectrum (higher entropy), which better tracks client utility.


\noindent
The choice of using the last layer for entropy computation motivates the use of Class-wise Shapley Values (CSSV) as an auxiliary signal for fusion. As discussed in subsection \ref{subsec:cssv}, CSSV is also computed on the last layer. Therefore, both the metrics used as inputs for fusion are derived from the same sub-part of the model, albeit with different approaches. Additionally, computing entropy on the last layer is the most efficient, as the Gram matrix $A \in \mathbb{R}^{C \times C}$ (see subsection \ref{subsec:entropy}) and scales only with the number of classes. 


\section{Compute Complexity}
\label{app:compute}
\noindent
For $n$ participating clients, let the final-layer update of each client be $M_i \in \mathbb{R}^{d \times C}$, where $d$ is the feature dimension and $C$ is the number of classes, with typically $d > C$. For each client, we first form the class-space Gram matrix, 
$A_i = M_i^\top M_i \in \mathbb{R}^{C \times C}$,
which costs $O(d\,C^2)$. We then compute the eigenspectrum of $A_i$ using SVD, which costs $O(C^3)$, and evaluate the entropy from the normalized eigenvalues in $O(C)$ time. Therefore, the total time complexity of entropy computation over all participating clients is
\begin{equation}
O\!\left(n\,d\,C^2 + n\,C^3\right).
\end{equation}
Because these computations
are performed independently per client, they are embarrassingly parallel across clients. 
The additional cost of \textsc{SpectralFuse} relative to \textsc{SpectralFed} comes from computing CSSV and applying the Kalman filter.

\noindent
We report the per-round \emph{server-side} wall-clock time for the baselines and our methods in Table~\ref{tab:runtime}.
The reported values are mean $\pm$ standard deviation in milliseconds over 10 rounds with all methods evaluated under identical CPU/GPU settings.

\begin{table}[!h]
\centering
\caption{Per-round server-side wall-clock time (ms) for running contribution estimation and aggregation.}
\label{tab:runtime}
\scriptsize
\begin{tabular}{lcccc}
\toprule
\textbf{Dataset} & \textbf{CGSV} & \textbf{ShapFed} & \textbf{SpectralFed} & \textbf{SpectralFuse} \\
\midrule
CIFAR-100 & 50 $\pm$ 2 & 467 $\pm$ 31 & 207 $\pm$ 2 & 378 $\pm$ 12 \\
FedISIC & 172 $\pm$ 6 & 398 $\pm$ 49 & 53 $\pm$ 10 & 102 $\pm$ 5 \\
\bottomrule
\end{tabular}
\end{table}

\noindent
Table~\ref{tab:runtime} shows that \textsc{SpectralFed} is substantially cheaper than ShapFed in both settings.
The entropy-based method are the fastest on FedISIC as they scale only with the number of classes. In contrast, the time complexity of CGSV scales with the full model size resulting in slower runtimes for larger models.
\end{document}